\RequirePackage{fix-cm}
\documentclass[smallextended]{svjour3}       
\smartqed  
\usepackage[text={7in,9in},centering]{geometry}
\usepackage{lineno}
\modulolinenumbers[5]
\usepackage{hyperref}

\usepackage{graphicx} 
\graphicspath{{./}}

\usepackage{caption}
\usepackage{multirow}
\usepackage{subcaption} 
\usepackage{booktabs} 
\usepackage{amsmath} 
\usepackage{color}
\usepackage[table]{xcolor}

\journalname{Integrating Materials and Manufacturing Innovation}
\begin{document}

\title{A comparative study of feature selection methods for stress hotspot classification in materials}
\thanks{United States National Science Foundation award number DMR-1307138 and DMR-1507830
}


\author{Ankita Mangal         \and
        Elizabeth A. Holm 
}


\institute{Ankita Mangal \at
              \email{mangalanks@gmail.com}           
           \and
           Elizabeth A. Holm \at
           Department of Materials Science and Engineering,\\
           Carnegie Mellon University\\
           5000 Forbes Ave\\
           Pittsburgh, PA 15213, USA\\
           \email{eaholm@andrew.cmu.edu}   
}

\date{Received: date / Accepted: date}

\maketitle

\begin{abstract}
The first step in constructing a machine learning model is defining the features of the data set that can be used for optimal learning. In this work we discuss feature selection methods, which can be used to build better models, as well as achieve model interpretability. We applied these methods in the context of stress hotspot classification problem, to determine what microstructural characteristics can cause stress to build up in certain grains during uniaxial tensile deformation. The results show how some feature selection techniques are biased and demonstrate a preferred technique to get feature rankings for physical interpretations. 
\end{abstract}

\section{Introduction}
\label{intro}
Statistical learning methods are gaining popularity in the materials science field, rapidly becoming known as "Materials Data Science". With new data infrastructure platforms like Citrination \cite{OMara2016} and the Materials data curation system \cite{Dima2016}, machine learning (ML) methods are entering the mainstream of materials science. Materials data science and informatics is an emergent field aligned with the goals of the Materials Genome Initiative to reduce the cost and time for materials design, development and deployment. Building and interpreting machine learning models are indispensable parts of the process of curating materials knowledge. ML methods have been used for predicting a target property such as material failure \cite{Mangal2017b,Mangal2017c}, twinning deformation \cite{Orme2016}, phase diagrams \cite{ChNg2017} and guiding experiments and calculations in composition space \cite{Ling2017,Oliynyk2016}. Machine learning models are built on learning from "features" or variables that describe the problem. Thus, an important aspect of the machine learning process is to determine which variables most enable data driven insights about the problem.

Dimensionality reduction techniques (such as principal component analysis(PCA) \cite{wall2003singular}, kernel PCA \cite{mika1999kernel}, autoencoders \cite{Holden2006}, feature compression from information gain theory \cite{yu2003feature}) have become popular for producing compact feature representations \cite{Guyon2003}. They are applied to the feature set to get the best feature representation, resulting in a smaller dataset, which speeds up the model construction \cite{van2009dimensionality}. However, dimensionality reduction techniques change the original representation of the features, and hence offer limited interpretability \cite{Guyon2003}. An alternate method for better models is feature selection. Feature selection is the process of selecting a subset of the original variables such that a model built on data containing only these features has the best performance. Feature selection avoids overfitting, improves model performance by getting rid of redundant features and has the added advantage of keeping the original feature representation, thus offering better interpretability \cite{Guyon2003}.

Feature selection methods have been used extensively in the field of bioinformatics \cite{Saeys2007}, psychiatry \cite{Lu2014} and cheminformatics \cite{Wegner2004}. There are multiple feature selection methods, broadly categorized into Filter, Wrapper and Embedded methods based on their interaction with the predictor during the selection process. The filter methods rank the variables as a preprocessing step, and feature selection is done before choosing the model. In the wrapper approach, nested subsets of variables are tested to select the optimal subset that work best for the model during the learning process. Embedded methods are those which incorporate variable selection in the training algorithm. 

We have used random forest models to study stress hotspot classification in FCC \cite{Mangal2017b} and HCP \cite{Mangal2017c} materials. In this paper, we review some feature selection techniques applied to the stress hotspot prediction problem in hexagonal close packed materials, and compare them with respect to future data prediction. We focus on two commonly used techniques from each method: (1) Filter Methods: Correlation based feature selection (CFS) \cite{Hall1999}, and Pearson Correlation \cite{Cohen2009}; (2) Wrapper Methods: Fealect \cite{Zare2013} and Recursive feature elimination (RFE) \cite{Guyon2003} and (3) Embedded Methods: Random Forest Permutation accuracy importance (RF-PAI) \cite{Breiman1996} and Least Absolute Shrinkage and Selection Operator (LASSO) \cite{Tibshirani1996}. The main contribution of this article is to raise awareness in the materials data science community about how different feature selection techniques can lead to misguided model interpretations and how to avoid them. We point out some of the inadequacies of popular feature selection methods and finally, we extract data driven insights with better understanding of the methods used.

\section{Methods}
\label{sec:1}
An applied stress is distributed heterogenously within the grains in a microstructure\cite{Qidwai2009}. Under an applied deformation, some grains are prone to accumulating stress due to their orientation, geometry and placement with respect to the neighboring grains. These regions of high stress, so called stress hotspots, are related to void nucleation under ductile fracture \cite{Rimmer1959}. Stress hotspot formation has been studied in face centered cubic (FCC)  \cite{Mangal2017b} and hexagonal close packed (HCP)  \cite{Mangal2017c} materials using a machine learning approach. A set of microstructural descriptors was  designed to be used as features in a random forest model for predicting stress hotspots. To achieve data driven insights into the problem, it is essential to rank the microstructural descriptors (features). In this paper, we review different feature selection techniques applied to the stress hotspot classification problem in HCP materials, which have a complex plasticity landscape due to anisotropic slip system activity.

Let $(x_i,y_i)$, for $i=1,...,N$ be N independent identically distributed (i.i.d.) observations of a p-dimensional vector of grain features $x_i \in R^p$, and the response variable $y_i \in {0,1}$ denotes the truth value of a grain being a stress hotspot. The input matrix is denoted by $X=(x_1,...,x_N)\in R^{N\times p}$, and $y\in[0,1]^N$ is the binary outcome. We will use small letters to refer to the samples $x_1,...,x_N$  and capital letters to refer to the features  $X_1,...,X_p$ of the input matrix $X$. Feature importance refers to metrics used by various feature selection methods to rank, such as feature weights in linear models or variable importance in random forest models.

\subsection{Dataset Studied}
A dataset of HCP microstructures with different textures was generated using Dream.3D in \cite{Mangal2017c}. Uniaxial tensile deformation was simulated in these microstructures using EVPFFT \cite{Lebensohn2012} with different constitutive parameters resulting in a dataset representing a Titanium like HCP material with an anisotropic critically resolved shear stress ratio \cite{Mangal2017c}. This dataset contains grain-wise values for equivalent Von Mises stress, and the corresponding Euler angles and grain connectivity parameters.

The grains having stress greater than the $90^{th}$ percentile of the stress distribution were designated as stress hotspots, a binary target. Thirty four variables to be used as features in machine learning were developed. These features ($X$) describe the grain texture and geometry and have been summarized in table \ref{Featurenames}. We rank these features using different feature selection techniques, and observe the improvement in models, as well as understand the physics behind stress hotspot formation. The model performance is measured by the AUC (area under curve), a metric for binary classification which is insensitive to imbalance in the classes. An AUC of 100\% denotes perfect classification and 50\% denotes no better than random guessing \cite{auc}.

\begin{table*}[t]
\centering
\caption{Feature name descriptions}
\label{Featurenames}
\small
\begin{tabular}{|p{0.18\linewidth}p{0.3\linewidth}|p{0.18\linewidth}p{0.3\linewidth}|}
\toprule 
\textbf{Feature name Abbreviation} & \textbf{Description} & \textbf{Feature name Abbreviation} & \textbf{Description}\\ \midrule
Schmid\_1 & Basal $<a>$ Schmid factor & 100\_IPF\_x & Distance of tensile axis from the corners of the 100 Inverse pole figure\\
\midrule
Schmid\_2 & Prismatic $<a>$ Schmid factor & 001\_IPF\_x & Distance of tensile axis from the corners of the 001 Inverse pole figure\\
\midrule
Schmid\_3 & Pyramidal $<a>$ Schmid factor & AvgC\_Axes\_x & Unit vector components describing the c axis orientation for hcp\\
\midrule
Schmid\_4 & Pyramidal $<c+a>$ Schmid factor & Max\_mis & Maximum misorientation between a grain and its nearest neighbor\\
\midrule
Schmid & FCC Schmid factor & Min\_mis & Minimum misorientation between a grain and its nearest neighbor\\
\midrule
$\theta$ & Polar angle of hcp c axis w.r.t sample frame & AvgMisorientations & Average misorientation between a grain and its nearest neighbor\\
\midrule
$\phi$ & Azimuthal Angle of hcp c axis w.r.t. sample frame & QPEuc & Average distance of a grain to quadruple junctions\\
\midrule
TJEuc & Average distance of a grain to triple junctions & NumNeighbors & Number of nearest neighbors of a grain\\
\midrule
GBEuc & Average distance of a grain to grain boundaries & Neighborhoods & Number of grains having their centroid within the 1 multiple of equivalent sphere diameters from each grain\\
\midrule
KernelAvg & Average misorientation within a grain & FeatureVolumes & Volume of grain\\
\midrule
Omega3s & 3rd invariant of the second-order moment matrix for the grain, without assuming a shape type & Equivalent Diameters & Equivalent spherical diameter of a grain\\
\midrule
mPrimeList & Slip transmission factor for fcc materials & AspectRatios & Ratio of axis lengths (b\/a and c\/a) for best-fit ellipsoid to grain shape\\
\midrule
Surface Features & 1 if grain touches the periodic boundary else 0 & Surface area volume ratio & Ratio between surface area and volume of a grain\\
\bottomrule
\end{tabular}
\end{table*}


\subsection{Feature Selection Methods}
\subsubsection{Filter Methods}
Filter methods are based on preprocessing the dataset to extract the features $X_1,...,X_p$ that most impact the target $Y$. Some of these methods are:

\paragraph{Pearson Correlation \cite{Cohen2009}:} This method provides a straightforward way for filtering features according to their correlation coefficient. The Pearson correlation coefficient between a feature $X_i$ and the target $Y$ is: $$\rho_i = \frac{cov(X_i, Y)}{\sigma_(X_i)\sigma_Y}$$ where $cov(X_i, Y)$ is the covariance, $\sigma$ is the standard deviation \cite{Cohen2009}. It ranges between $(-1,1)$ from negative to positive correlation, and can be used for binary classification and regression problems. It is a quick metric using which the features are ranked in order of the  absolute correlation coefficient to the target. 

\paragraph{Correlation based feature selection (CFS) \cite{Hall1999}:} CFS was developed to select a subset of features with high correlation to the target and low intercorrelation among themselves, thus reducing redundancy and selecting a diverse feature set. CFS gives a heuristic merit over a feature subset instead of individual features. It uses symmetrical uncertainty correlation coefficient given by: $$ r(X,Y) = 2.0 \times \frac{IG(X|Y)}{H(X)+H(Y)} $$ where $IG(X|Y)$ is the information gain of feature $X$ for the class attribute $Y$. $H(X)$ is the entropy of variable $X$. The following merit metric was used to rank each subset $S$ containing $k$ features: $$Merit_S = \frac{k\overline{r_{cf}}}{\sqrt{k + k(k-1)\overline{r_{ff}}}}$$ 
where $\overline{r_{cf}}$ is the mean symmetrical uncertainty correlation between the feature $(f\in S)$ and the target, and $\overline{r_{ff}}$ is the average feature-feature inter-correlation. To account for the high computational complexity of evaluating all possible feature subsets, CFS is often combined with search strategies such as forward selection, backward elimination and bi-directional search. In this work we have used the scikit-learn implementation of CFS \cite{Zhao2010} which uses symmetrical uncertainity \cite{Hall1999} as the correlation metric and explores the subset space using best first search \cite{Pearl:1984:HIS:525}, stopping when it encounters five consecutive fully expanded non-improving subsets.

\subsubsection{Embedded Methods}
These methods are popular because they perform feature selection while constructing the classifier, removing the preprocessing feature selection step. Some popular algorithms are support vector machines (SVM) using recursive feature elimination (RFE) \cite{Guyon2002}, random forests (RF) \cite{Breiman1996} and Least absolute shrinkage and selection operator (LASSO)\cite{Tibshirani1996}. We compare LASSO and RF methods for feature selection on the stress hotspot dataset.

\paragraph{Least Absolute Shrinkage and Selection Operator (LASSO) \cite{Tibshirani1996}:}
LASSO is linear regression with $L_1$ regularization \cite{Tibshirani1996}.  A linear model $\mathcal{L}$ is constructed  $$\mathcal{L}: min_{w\in R^p}\sum_{i=1}^{N}\frac{1}{2N}||y_i - w^T\cdot x_i||_2^2 + \lambda||w||_1$$ on the training data $(x_i, y_i)$, $i=1....,N$, where $w$ is a $p$ dimensional vector of weights corresponding to each feature dimension $p$. The $L_1$ regularization term ($ \lambda||w||_1$) helps in feature selection by pushing the weights of correlated features to zero, thus preventing overfitting and improving model performance. Model interpretation is possible by ranking the features according to the LASSO feature weights. However, it has been shown that for a given regularization strength $\lambda$, if the features have redundancy, inconsistent subsets can be selected \cite{bach2008bolasso}. Nonetheless, Lasso has been shown to provide good prediction accuracy by reducing model variance without substantially increasing the bias while providing better model interpretability. We used the scikit-learn implementation to compute our results \cite{pedregosa2011scikit}.

\paragraph{Random Forest Permutation Accuracy importance (RF PAI) \cite{Breiman1996}:}
The random forest is a non linear multivariate model built on an ensemble of decision trees. It can be used to determine feature importance using the inbuilt feature importance measure \cite{Breiman1996}. For each of the trees in the model, a feature node is randomly replaced with another feature node while keeping all others nodes unchanged. The resulting model will have a lower performance if the feature is important. When the permuted variable $X_j$, together with the remaining unchanged variables, is used to predict the response, the number of observations classified correctly decreases substantially, if the original variable $X_j$ was associated with the response. Thus, a reasonable measure for feature importance is the difference in prediction accuracy before and after permuting $X_j$. The feature importance calculated this way is known as Permutation Accuracy Importance (PAI) and was computed using the scikit-learn package in Python \cite{pedregosa2011scikit}. 

\subsubsection{Wrapper Methods}
Wrapper methods test feature subsets using a model hypothesis. Wrapper methods can detect feature dependencies i.e. features that become importance in presence of each other. They are computationally expensive, hence often use greedy search strategies (forward selection and backward elimination \cite{sutter1993comparison}) which are fast and avoid overfitting to get the best nested subset of features.

\paragraph{Fealect Algorithm \cite{Zare2013}:}
The number of features selected by Lasso depends on the regularization parameter $\lambda$, and in the presence of highly correlated features, LASSO arbitrarily selects one feature from a group of correlated features \cite{Zou2005}. The set of possible solutions for all LASSO regularization strengths is given by the regularization path, which can be recovered computationally efficiently using the Least Angles Regression (LARS) algorithm \cite{Efron2004}. It was shown that LASSO selects the the relevant variables with a probability one and all other with a positive probability \cite{bach2008bolasso}. An improvement in LASSO, the Bolasso feature selection algorithm was developed based on this property \cite{bach2008bolasso} in 2008. In this method, the dataset is bootstrapped, and a LASSO model with a fixed regularization strength $\lambda$ is fit to each subset. Finally, the intersection of the LASSO selected features in each subset is chosen to get a consistent feature subset.

In 2013, the FeaLect algorithm, an improvement over the Bolasso algorithm, was developed based on the combinatorial analysis of regression coefficients estimated using LARS \cite{Zare2013}. FeaLect considers the full regularization path, and computes the feature importance using a combinatorial scoring method, as opposed to simply taking the intersection with Bolasso. The FeaLect scoring scheme measures the quality of each feature in each bootstrapped sample, and averages them to select the most relevant features, providing a robust feature selection method. We used the R implementation of FeaLect to compute our results \cite{Zare2015}.

\paragraph{Recursive Feature Elimination (RFE) \cite{Guyon2002}:}
A number of common ML techniques (such as linear regression, support vector machines (SVM), decision trees, Naive Bayes, perceptron, e.t.c) provide feature weights that consider multivariate interacting effects between features \cite{Guyon2003}. To interpret the relative importance of the variables from these model feature weights, RFE was introduced in the context of support vector machines (SVM) \cite{Guyon2002} for getting compact gene subsets from DNA-microarray data. 

To find the best feature subset, instead of doing an exhaustive search over all feature combinations, RFE uses a greedy approach, which has been shown to reduce the effect of correlation bias in variable importance measures \cite{Gregorutti2016}. RFE uses backward elimination by taking the given model (SVM, random forests, linear regression etc.) and discarding the worst feature (by absolute classifier weight or feature ranking), and repeating the process over increasingly smaller feature subsets until the best model hypothesis is achieved. The weights of this optimal model are used to rank features. Although this feature ranking might not be the optimal ranking for individual features, it is often used as a variable importance measure \cite{Gregorutti2016}. We used the scikit-learn implementation of RFE with random forest classifier to come up with a feature ranking for our dataset.

\section{Results and Discussion}

\begin{table*}[!htbp]
\begin{center}
\caption{Variable Importance Measures using different methods for HCP materials with Unequal CRSS. The gray shaded cells denote the features selected by the corresponding technique. The features describing grain geometry are shaded in green.}
\label{VarImp_UnequalCRSS}
\begin{tabular}{p{3cm}* {8}{p{1.3cm}}}
\toprule 

\textbf{Features} & \textbf{Pearson Correlation} & \textbf{CFS} & \textbf{RFE} & \textbf{RF ($\times 1e-2$)} & \textbf{Linear Regression} & \textbf{Ridge Regression} & \textbf{LASSO Regression} & \textbf{Fealect ($\times 1e-2$)}\\ 

\midrule

$cos\phi$ & \cellcolor{lightgray}-0.29 & \cellcolor{lightgray}1 & \cellcolor{lightgray}1 & \cellcolor{lightgray}53.43 & \cellcolor{lightgray}27.37 & \cellcolor{lightgray}27.36 & \cellcolor{lightgray}26.01 & \cellcolor{lightgray}245.0\\

$Schmid\_1$ & \cellcolor{lightgray}-0.39 & 0 & \cellcolor{lightgray}1 & \cellcolor{lightgray}0.15 & \cellcolor{lightgray}22.72 & \cellcolor{lightgray}22.69 & \cellcolor{lightgray}14.78 & \cellcolor{lightgray}145.00\\

\cellcolor{green}EquivalentDiameters & -0.01 & 0 & \cellcolor{lightgray}1 & 0.05 & 0.15 & 0.15 & \cellcolor{lightgray}0.08 & \cellcolor{lightgray}79.47\\

\cellcolor{green}GBEuc & -0.01 & 0  &  \cellcolor{lightgray}1 & \cellcolor{lightgray}0.12 & 0.22 & 0.22 & \cellcolor{lightgray}0.12 & \cellcolor{lightgray}71.47\\

$Schmid\_4$ & \cellcolor{lightgray}-0.18 & 0 & \cellcolor{lightgray}1 & \cellcolor{lightgray} 0.31 & \cellcolor{lightgray}7.29 & \cellcolor{lightgray}7.31 & \cellcolor{lightgray}10.35 & \cellcolor{lightgray}41.27\\

\cellcolor{green}Neighborhoods & -0.01 & 0 & 22 & 0.01 & 0.10 & 0.10 & 0.00 & \cellcolor{lightgray}5.53\\

$sin\theta$ & \cellcolor{lightgray}0.48 & \cellcolor{lightgray}1 & \cellcolor{lightgray}1 & \cellcolor{lightgray}8.74 & \cellcolor{lightgray}74.78 & \cellcolor{lightgray}74.61 & \cellcolor{lightgray}52.99 & \cellcolor{lightgray}5.00\\

\cellcolor{green}TJEuc & -0.01 & 0 & 2 & 0.07 & 0.97 & \cellcolor{lightgray}0.97 & \cellcolor{lightgray}0.44 & \cellcolor{lightgray}4.93\\

$sin\phi$ & \cellcolor{lightgray}0.14 & \cellcolor{lightgray}1 & 16 & 0.03 & \cellcolor{lightgray}80.46 & \cellcolor{lightgray}79.96 & \cellcolor{lightgray}19.17 & \cellcolor{lightgray}1.0\\

AvgMisorientations & \cellcolor{lightgray}0.31 & 0 & \cellcolor{lightgray}1 & \cellcolor{lightgray} 8.95 & \cellcolor{lightgray}32.08 & \cellcolor{lightgray}32.09 & \cellcolor{lightgray}32.05 & \cellcolor{lightgray} 0.83\\

\cellcolor{green}NumNeighbors & -0.01 & 0 & 23 & 0.01 & 0.18 & 0.17 & \cellcolor{lightgray}0.03 & 0.50 \\

$Schmid\_3$ & \cellcolor{lightgray}0.12 & 0 & 9 & 0.03 & \cellcolor{lightgray}4.05 & 4.04 & 0.00 & 0.0\\

$Min\_mis$ & \cellcolor{lightgray}0.09 & 0 & \cellcolor{lightgray}1 & \cellcolor{lightgray}0.72 & \cellcolor{lightgray}3.46 & \cellcolor{lightgray}3.46 & \cellcolor{lightgray}2.19 &  0.0\\

$AvgC\_Axes\_1$ & 0.00 & 0 & \cellcolor{lightgray}1 & \cellcolor{lightgray}0.22 & 0.09 & 0.09 & 0.00 & 0.0\\

$Max\_mis$ & \cellcolor{lightgray}0.17 & 0 & 4 & 0.02 & 0.86 & 0.86 & \cellcolor{lightgray}0.03 & 0.0\\

\cellcolor{green}NumCells & -0.01 & 0 & 18 & 0.04 & \cellcolor{lightgray}1.3e6 & 0.11 & \cellcolor{lightgray}0.21 & 0.0\\

$Schmid\_2$ & \cellcolor{lightgray}0.49 & 0 & \cellcolor{lightgray}1 & \cellcolor{lightgray}26.80 & \cellcolor{lightgray}38.03 & \cellcolor{lightgray}37.83 & \cellcolor{lightgray}8.37 & 0.0\\

KernelAvg & -0.01 & 0 & 25 & 0.0 & 0.22 & 0.22 & 0.00 & 0.0\\

$010\_IPF\_1$ & \cellcolor{lightgray}-0.07 & 0 & 5 & 0.01 & 0.49 & 0.49 & 0.00 & 0.0\\

$\phi$ & \cellcolor{lightgray}0.13 & \cellcolor{lightgray}1 & 3 & \cellcolor{lightgray}3.4 & \cellcolor{lightgray}66.42 & \cellcolor{lightgray}65.94 & \cellcolor{lightgray}7.68 & 0.0\\

$001\_IPF\_0$ & 0.00 & 0 & 11 & 0.03 & 0.58 & 0.57 & 0.00 & 0.0\\

$001\_IPF\_2$ & \cellcolor{lightgray}0.09 & 0 & 21 & 0.01 & 0.21 & 0.24 & \cellcolor{lightgray}0.19 & 0.0\\

$010\_IPF\_0$ & 0.00 & 0 & 12 & 0.01 & 0.76 & 0.76 & \cellcolor{lightgray}0.23 & 0.0\\

$100\_IPF\_0$ & 0.00 & 0 & 10 & 0.01 & 0.13 & 0.13 & 0.00 & 0.0\\

$001\_IPF\_1$ & \cellcolor{lightgray}0.16 & 0 & 15 & 0.01 & 0.17 & 0.14 & 0 & 0.0\\

$100\_IPF\_1$ & \cellcolor{lightgray}0.07 & 0 & 14 & 0.02 & \cellcolor{lightgray}1.10 & \cellcolor{lightgray}1.10 & 0.00 & 0.0\\

\cellcolor{green}QPEuc & -0.01 & 0 & 6 & 0.02 & 0.57 & 0.57 & 0.00 & 0.0\\

$AvgC\_Axes\_0$ & 0.00 & 0 & 7 & 0.03 & 0.34 & 0.34 & \cellcolor{lightgray}0.05 & 0.0\\

$\theta$ & 0.00 & \cellcolor{lightgray}1 & 24 & 0.02 & 0.04 & 0.04 & 0.00 & 0.0\\

\cellcolor{green}FeatureVolumes & -0.01 & 0 & 13 & 0.04 & \cellcolor{lightgray}1.3e6 & 0.11 & 0.00 & 0.0\\

$010\_IPF\_2$ & -0.04 & 0 & 17 & 0.01 & 0.79 & 0.79 & 0.00 & 0.0\\

$AvgC\_Axes\_2$ & 0.00 & 0 & 8 & 0.01 & \cellcolor{lightgray}2.9e4 & 0.07 & 0.00 & 0.0\\

$100\_IPF\_2$ & 0.04 & 0 & 19 & 0.01 & \cellcolor{lightgray}1.21 & \cellcolor{lightgray}1.20 & 0.00 & 0.0\\

$cos\theta$ & 0.00 & \cellcolor{lightgray}1 & 20 & 0.01 & \cellcolor{lightgray}2.9e4 & 0.07 & 0.00 & 0.0\\
\midrule
\multicolumn{9}{c}{Random Forest model AUC without feature selection: 71.94\% }\\
\midrule
\multicolumn{9}{c}{Random Forest model AUC with selected features (\%) }\\
\midrule
training & 84.02 & 82.51 & 84.24 & 83.82 & 84.20 & 84.19 & 84.31 & 84.28\\
validation & 80.46 & 80.45 & 80.73 & 80.19 & 80.72 & 80.61 & 80.83 & 80.75\\
\bottomrule
\end{tabular}
\end{center}
\end{table*}

\begin{figure*}[!h]
\centering
\includegraphics[width = \textwidth]{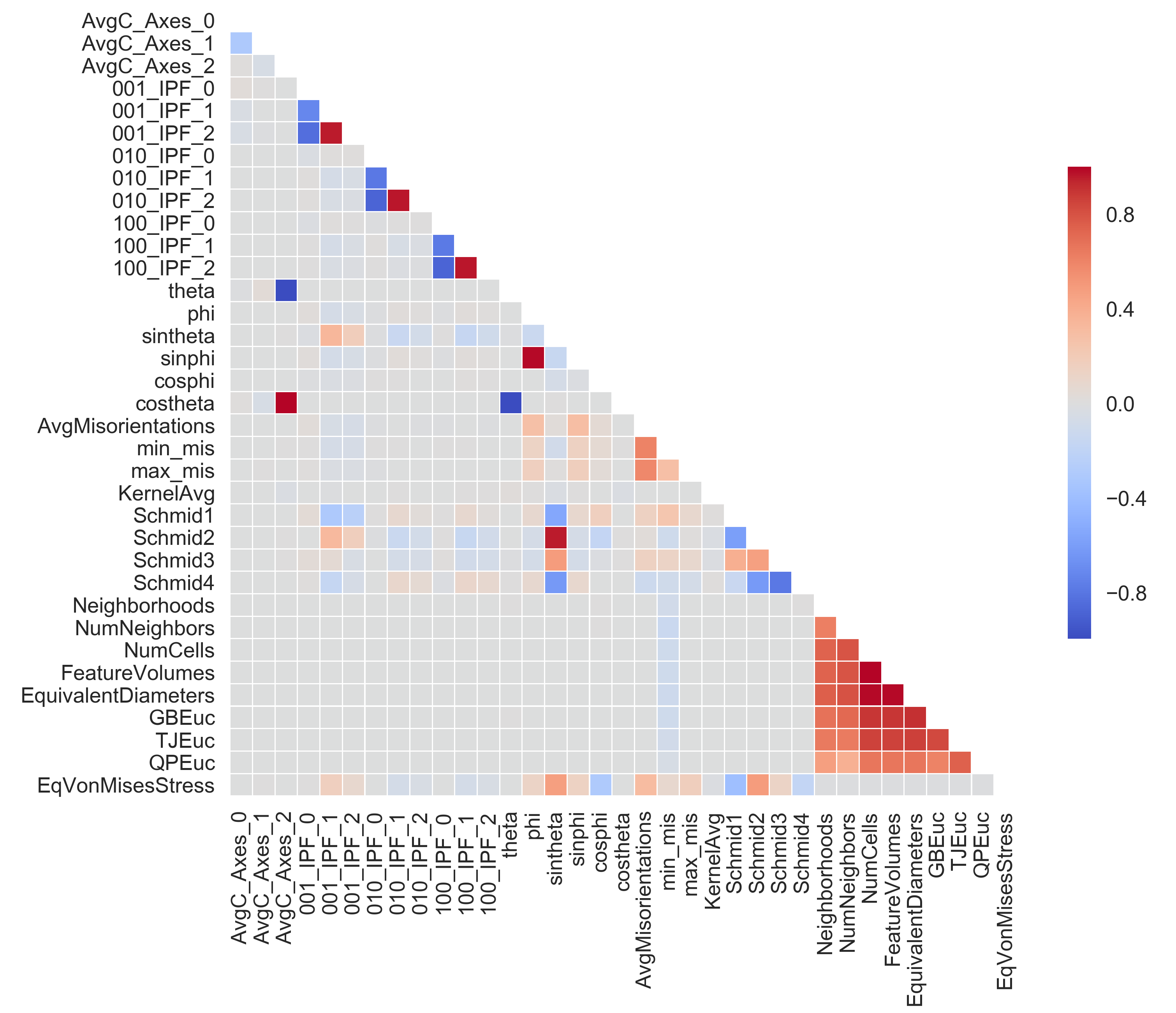}
    \caption{Pearson Correlation matrix between the target (EqVonMisesStress) and all the features}
    \label{fig:FeatureCorrelation}
\end{figure*}

Table \ref{VarImp_UnequalCRSS} shows the feature importances calculated using filter based methods: Pearson correlation and CFS; embedded methods: Random Forest (RF), Linear regression, Ridge regression ($L_2$ regularization) and LASSO regression and finally wrapper methods: RFE and Fealect . The shaded cells denote the features that were finally selected to build RF models and their corresponding performances are noted. The input data was scaled by minimum and maximum values to [0,1]. Figure \ref{fig:FeatureCorrelation} shows the correlation matrix for the features and the target.

Pearson correlation can be used for feature selection, resulting in a good model. However, this measure has implicit orthogonality assumptions between variables, and the coefficient does not take mutual information between features into account. Additionally, this method only looks for linear correlations which might not capture many physical phenomenon.

The feature subset selected by CFS contains features with higher class correlation and lower redundancy, which translate to a good predictive model. Although we know grain geometry and neighborhood are important to hotspot formation, CFS does not select any geometry based features and fails to provide an individual feature ranking.

Linear regression, ridge regression and Lasso are highly correlated linear models. A simple linear model results in huge weights for some features (NumCells, FeatureVolumes), likely due to overfitting, and hence is unsuitable for deducing variable importance. Ridge regression compensates for this problem by using $L_1$ regularization, but the weights are distributed among the redundant features, which might lead to incorrect conclusions. LASSO regression overcomes this problem by pushing the weights of correlated features to zero, resulting in a good feature subset. The top five ranked features by LASSO with regularization strength of $\lambda =0.3$ are : $sin\theta$, AvgMisorientations, $cos\phi$, $sin\phi$ and $Schmid\_1$. The first geometry based feature ranks $10^{th}$ on the list, which seems to underestimate the physical importance of such features. A drawback of deriving insights from LASSO selected features is that it arbitrarily selects a few representatives from the correlated features, and the number of features selected depends heavily on the regularization strength. Thus the models become unstable, because changes in training subset can result in different selected features. Hence these methods are not ideal for deriving physical insights from the model.

Random forest models also provide an embedded feature ranking module. The RF-PAI importance seems to focus only on the hcp 'c' axis orientation derived features ($cos\phi, sin\theta, $), average misorientation and the Prismatic $<a>$ Schmid factor, while discounting most of the geometry derived features. RF-PAI suffers from correlation bias due to preferential selection of correlated features during tree building process \cite{Strobl2008}. As the number of correlated variables increases, the feature importance score for each variable decreases. Often times the less relevant variables replace the predictive ones (due to correlation) and thus receive undeserved, boosted importance \cite{Tolosi2011}. Random forest variable importance can also be biased in situations where the features vary in their scale of measurement or number of categories, because the underlying Gini gain splitting criterion is a biased estimator and can be affected by multiple testing effects \cite{Strobl2007}. From Figure \ref{fig:FeatureCorrelation}, we found that all the geometry based features are highly correlated to each other, therefore deducing physical insights from this ranking is unsuitable. 

Hence, we move to Wrapper based methods for feature importance. Recursive feature elimination (RFE) has been shown to reduce the effect of the correlation on the importance measure \cite{Gregorutti2016}. RFE with underlying random forest model selects a feature subset consisting of two geometry based features (GBEuc and EquivalentDiameter), however, it fails to give an individual ranking among the features.

FeaLect provides a robust feature selection method by compensating for the uncertainty in LASSO due to arbitrary selection among correlated variables, and the number of selected variables due to change in regularization strength. Table \ref{VarImp_UnequalCRSS} lists the Fealect selected variables in decreasing order. We find that the top two important features are derived from the grain crystallography, and geometry derived features come next. This suggests that both texture and geometry based features are important. Using linear regression based methods such as these tell us which features are important by themselves, as opposed to RF-PAI which indicates the features that become important due to interactions between them (via RF models) \cite{Guyon2003}. The Fealect method provides the best estimate of the feature importance ranking which can then be used to extract physical insights. This method also divides the features into 3 classes: informative, irrelevant features that cause model overfitting and redundant features \cite{Zare2013}. The most informative features are: $cos\phi$, $Schmid\_1$, EquivalentDiameter, GBEuc, $Schmid\_4$, Neighborhoods, $sin\theta$ and TJEuc. The irrelevant features are $sin\phi$ and AvgMisorientations (which cause model overfitting). The remaining features are redundant. 

A number of selected features directly or indirectly represent the HCP c-axis orientation, such as $cos\phi$,  $sin\theta$ and basal Schmid factor ($Schmid\_1$), which is proportional to $cos\theta$. It is interesting that pyramidal $< c+a >$ Schmid factor ($Schmid\_4$) is chosen as important. From Figure \ref{fig:FeatureCorrelation}, we can see that hot grains form where $\theta, \phi$ maximize $sin\theta$ and  $sin\phi$ i.e. $\theta \sim 90,\phi \sim 90$. This means that the HCP c-axis orientation of hot grains aligns with the sample Y axis, which means these grains have a low elastic modulus. Since the c-axis is perpendicular to the tensile axis (sample Z); the deformation along the tensile direction can be accommodated by prismatic slip in these grains, and if pyramidal slip is occurring, it means they have a very high stress \cite{Mangal2017c}. This explains the high importance of the pyramidal $< c+a >$ Schmid factor. From the Pearson correlation coefficients in Figure \ref{fig:FeatureCorrelation}, we can observe that the stress hotspots form in grains with low basal and pyramidal $< c + a >$ Schmid factor, high prismatic $< a >$ Schmid factor, and higher values of $sin\theta$ and $sin\phi$. 

From Figure \ref{fig:FeatureCorrelation}, we can see that all the grain geometry descriptors do not have a direct correlation with stress, but are still selected by Fealect. This points to the fact that these variables become important in association with others. We analyzed these features in detail in \cite{Mangal2017c} and found that the hotspots lie closer to grain boundaries (GBEuc), triple junctions (TJEuc), and quadruple points (QPEuc), and prefer to form in smaller grains.



There is a subtle distinction between the physical impact of a variable on the target vs. the variables that work best for a given model. From table \ref{VarImp_UnequalCRSS}, we can see that a random forest model built on the entire feature set without feature selection has an AUC of $71.94\%$. All the feature selection techniques result in an improvement in the performance of the random forest model to a validation AUC of about 81\%. However, to draw physical interpretations, it is important to use a feature selection technique which: 1) keeps the original representation of the features, 2) is not biased by correlations/ redundancies among features, 3) is insensitive to the scale of variable values , 4) is stable to the changes in the training dataset, 5) takes multivariate dependencies between the features into account, and 6) provides an individual feature ranking measure.

\section{Conclusions}
We have used different feature selection techniques and demonstrated that while all techniques lead to an improvement in model performance, only the FeaLect method helps us to determine the underlying importance of the features by themselves.
\begin{itemize}
 \item All feature selection techniques result in $\sim 9\%$ improvement in the AUC metric for stress hotspot classification.
 \item Correlation based feature selection and Recursive feature elimination are computationally expensive to run, and give only a feature subset ranking.
 \item Random forest embedded feature ranking is biased against correlated features and hence should not be used to derive physical insights.
 \item Linear regression based feature selection techniques can objectively denote the most important features, however have their flaws. The Fealect algorithm can compensate for the variability in LASSO regression, providing a robust feature ranking that can be used to derive insights.
 \item Stress hotspots formation under uniaxial tensile deformation is determined by a combination of crystallographic and geometric microstructural descriptors.
 \item It is essential to choose a feature selection method that can find this dependence even when features are redundant or correlated.
\end{itemize}

\begin{acknowledgements}
This work was performed at Carnegie Mellon University and has ben supported by the United States National Science Foundation award number DMR-1307138 and DMR-1507830. The authors are grateful to the authors of skfeature and sklearn python libraries who made their source code available through the Internet. We would also like to thank the reviewers for their thorough work.
\end{acknowledgements}


\bibliographystyle{spphys}       
\bibliography{library}

\end{document}